\documentclass{article}

\usepackage{arxiv}

\usepackage[utf8]{inputenc} 
\usepackage[T1]{fontenc}    
\usepackage{hyperref}       
\usepackage{url}            
\usepackage{booktabs}       
\usepackage{amsfonts}       
\usepackage{nicefrac}       
\usepackage{microtype}      
\usepackage{lipsum}		
\usepackage{graphicx}
\usepackage{doi}

\usepackage{algorithm}
\usepackage{algpseudocode}
\usepackage{amsmath}

\usepackage{multirow}

\title{SparseAccelerate: Efficient Long-Context Inference for Mid-Range GPUs}

\author{ \href{https://orcid.org/0000-0002-4363-2177}{\includegraphics[scale=0.06]{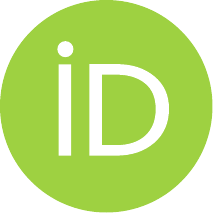}\hspace{1mm}James Vo}\thanks{Anh-Dung Vo} \\
	DocumentAI Team\\
	AGILESODA INC.\\
	Seoul, 06149, South Korea \\
	\texttt{anhdungitvn@agilesoda.ai} \\
}

\renewcommand{\shorttitle}{SparseAccelerate: Efficient Long-Context Inference for Mid-Range GPUs}

\hypersetup{
pdftitle={SparseAccelerate: Efficient Long-Context Inference for Mid-Range GPUs},
pdfsubject={q-bio.NC, q-bio.QM},
pdfauthor={David S.~Hippocampus, Elias D.~Striatum},
pdfkeywords={First keyword, Second keyword, More},
}

\begin{document}
\maketitle

\begin{abstract}

As Large Language Models (LLMs) scale to longer context windows, the computational cost of attention mechanisms, which traditionally grows quadratically with input length, presents a critical challenge for real-time and memory-constrained deployments. Existing sparse attention techniques have sought to reduce this complexity, but they often incur significant overhead or compromise accuracy, making them less practical for large contexts on mid-range hardware. In this paper, we introduce SparseAccelerate, a dynamic sparse attention method that adapts its sparsity patterns based on input characteristics, effectively flattening the attention complexity curve. Our approach is effective for input lengths starting at 16K tokens and scales efficiently up to 128K tokens on dual NVIDIA A5000 GPUs (24GB each). Experimental results show that SparseAccelerate achieves up to a 1.04x reduction in Time-To-First-Token (TTFT) latency at 32K tokens, while also providing substantial memory savings. These improvements yield practical gains for memory-intensive applications and long-context tasks that were previously infeasible with standard attention. Beyond latency reductions, SparseAccelerate fundamentally shifts the scaling trend, demonstrating the smallest TTFT growth gradient relative to context length among competing methods. Ongoing evaluations on diverse benchmarks confirm its scalability, positioning SparseAccelerate as a critical advancement toward efficient, real-time, and large-context LLM inference on accessible hardware.

\end{abstract}


\section{Introduction}
LLMs have demonstrated remarkable capabilities across a wide range of complex tasks, including code generation, long-form document comprehension, and advanced reasoning over extended contexts. Recent advancements, exemplified by models like LLaMA and GPT, have significantly expanded the scope of LLMs to handle context windows ranging from hundreds of thousands to even millions of tokens. This progress has unlocked new applications, such as Retrieval-Augmented Generation (RAG), long-form conversational AI, and context-aware question answering. However, these capabilities come with a critical limitation: the quadratic computational complexity of the attention mechanism.

As context lengths grow, the associated computational and memory costs escalate, leading to substantial delays in generating the first output token—a crucial metric for real-time and latency-sensitive applications. For example, an 8B-parameter LLM can take 10 to 20 seconds to produce the first token for a 32K-token input on dual NVIDIA A5000 GPUs (24GB each). This performance bottleneck renders such models impractical for deployment on mid-range hardware, particularly for applications that demand long-context processing or real-time responsiveness, such as context-aware question answering or RAG.

Numerous studies have been conducted to accelerate the inference process~\cite{agrawal2023sarathiefficientllminference, holmes2024deepspeedfastgenhighthroughputtextgeneration}. Previous research has attempted to tackle this challenge by utilizing sparse attention mechanisms, which reduce computational load through fixed sparsity patterns~\cite{jiang2023mistral7b, ding2023longnetscalingtransformers1000000000, lagunas2021blockpruning, nawrot2024dynamicmemorycompressionretrofitting, zimerman2024explainingmoderngatedlinearrnns, dao2024transformersssmsgeneralizedmodels, ribar2024sparqattentionbandwidthefficientllm}. While these methods offer some efficiency improvements, they fall short in adapting to the dynamic nature of attention distributions across different inputs. This limitation leads to suboptimal performance, especially in scenarios that require long-context understanding or high precision at scale.

To address this challenge, we introduce SparseAccelerate, a dynamic and adaptive method primarily aimed at accelerating the pre-filling phase of long-context LLM inference, with additional improvements in the decoding phase. SparseAccelerate leverages dynamically identified sparse attention patterns, which are adaptively determined at runtime based on the input sequence. This approach significantly reduces the computational cost of attention mechanisms, while maintaining or even improving model accuracy. In contrast to traditional sparse attention methods that rely on static patterns, our method optimizes sparsity dynamically, striking a balance between computational efficiency and model precision.

The key contributions of this work are:  

\begin{itemize}  
    \item \textbf{Dynamic Sparse Attention Patterns:}  
    We identify and exploit three generalized sparsity patterns—\textit{Triangular}, \textit{Interval-Slash}, and \textit{Block-Cluster}—that are both computationally efficient and effective for long-context attention.  

    \item \textbf{Kernel-Aware Optimization Framework:}  
    To minimize computational overhead, we develop a kernel-aware optimization framework that dynamically selects the optimal sparsity pattern for each attention head at runtime, maximizing hardware utilization.  

    \item \textbf{Significant Speedup with Scalability:}  
    SparseAccelerate achieves up to a 1.04x reduction in TTFT latency for 32K-token prompts on dual NVIDIA A5000 GPUs (24GB each). Furthermore, our method demonstrates the smallest TTFT gradient with respect to context length compared to state-of-the-art approaches, effectively breaking the quadratic scaling bottleneck.  
\end{itemize}  

SparseAccelerate provides a practical and scalable solution for deploying long-context LLMs on mid-range hardware, enabling real-time applications like retrieval-augmented generation (RAG), long-form document understanding, and context-aware question answering.

In the following sections, we provide a detailed explanation of our proposed method, followed by a description of the experimental setup used to evaluate its performance. We then present and analyze the experimental results, highlighting the effectiveness and scalability of our approach. Finally, we discuss potential avenues for future work to further enhance SparseAccelerate and address remaining challenges.

\section{Proposed Method}
The inference process consists of two distinct phases: the \textit{prefill phase}, which processes the input prompt, and the \textit{decode phase}, which generates output tokens in an autoregressive manner. In the prefill phase, all input tokens of a given batch are processed simultaneously. This is represented by a tensor \(X\) of shape \([B, L, H]\), where \(B\) is the batch size, \(L\) is the sequence length of each request, and \(H\) is the model's embedding size. In contrast, during the decode phase, the same operations are performed, but only for the single token generated in the last autoregressive iteration. Consequently, the input tensor in the decode phase has a shape of \([B, 1, H]\), as opposed to \([B, L, H]\) in the prefill phase.

The time required to generate the first token, known as \textit{Time To First Token} (TTFT), primarily reflects the runtime of the prefill phase, while the time between successive tokens, referred to as \textit{Inter Token Latency} (ITL), is mainly determined by the decoding process.

When processing long-context inputs, where \(L\) is significantly larger than 1, two key effects arise:
\begin{enumerate}
    \item TTFT becomes substantially larger than ITL. The prefill phase, which involves processing the entire input sequence, significantly increases the time required for generating the first token. 
    \item As a result, TTFT grows considerably, meaning that there is a prolonged waiting time before the first token is produced.
\end{enumerate}

This highlights the challenge of efficiently managing long-context inputs, as the increased prefill time can create a significant delay before the generation process begins. 

Most of the computational cost and runtime in this process lies within the attention module, due to the operations involved in the preprojection, attention (attn), and postprojection (postproj) stages. To address this, we experiment with reducing these costs by substituting a more suitable attention mechanism for the inference process, with the goal of improving overall efficiency while maintaining performance.

\subsection{Kernel-Aware Sparse Pattern Search Algorithm}
\textbf{Algorithm Idea:}  
This algorithm dynamically identifies the optimal sparse attention pattern by iteratively refining a search space of patterns. The goal is to balance computational efficiency and attention quality by ensuring that the computational cost (in FLOPs) closely matches a predefined target.

\textbf{Description:}  
The kernel-aware search iterates through potential patterns, measuring their computational cost and adjusting dynamically to align with the target FLOPs. This process ensures the most suitable sparse attention pattern is selected for each attention head.

\begin{algorithm}
\caption{Optimal sparse attention pattern selection by dynamically refining a search space based on target computational cost and accuracy.}
\label{alg:kernel_aware}
\begin{algorithmic}[1]
\Require Queries $\mathbf{Q}$, Keys $\mathbf{K}$, Values $\mathbf{V} \in \mathbb{R}^{N \times d_h}$; candidate patterns $\mathcal{P}$; search space $\mathcal{S}$; target FLOPs $t_{\text{target}}$; initialized search space $\mathcal{S}_0$
\Ensure Optimal sparse pattern $p_{\text{opt}}$
\State \textbf{Initialize Search Space:}
\For{each candidate $\mathcal{S}_i$ in $\mathcal{S}_0$}
    \State Compute FLOPs $t_i \gets \text{FLOPs}(\mathcal{S}_i)$
    \While{$|t_i - t_{\text{target}}| > \epsilon$}
        \State Adjust $\mathcal{S}_i$ using $\mathcal{P}$: $\mathcal{S}_i \gets \text{RefineSearchSpace}(\mathcal{S}_i, \mathcal{P})$
        \State Recompute FLOPs $t_i \gets \text{FLOPs}(\mathcal{S}_i)$
    \EndWhile
    \State Add $\mathcal{S}_i$ to the final search space: $\mathcal{S} \gets \mathcal{S} \cup \mathcal{S}_i$
\EndFor
\State \textbf{Evaluate Patterns:}
\State Compute full attention output $\mathbf{A}_{\text{dense}} \gets \text{Softmax}(\mathbf{QK}^\top / \sqrt{d})$
\State Initialize best pattern $p_{\text{opt}} \gets \emptyset$
\For{each search space $\mathcal{S}_i$ in $\mathcal{S}$}
    \State Sparse output $\mathbf{A}_{\text{sparse}} \gets \text{SparseAttention}(\mathbf{QK}^\top / \sqrt{d}, \mathcal{S}_i)$
    \State Update best pattern: $p_{\text{opt}} \gets \arg\min(\|\mathbf{A}_{\text{sparse}} - \mathbf{A}_{\text{dense}}\|, p_{\text{opt}})$
\EndFor
\State \Return $p_{\text{opt}}$
\end{algorithmic}
\end{algorithm}

\subsection{Vertical-Slash Sparse Attention Algorithm}
\textbf{Algorithm Idea:}  
This algorithm efficiently computes attention for the most relevant vertical and diagonal (slash) positions in the attention matrix, focusing on these regions while ignoring less relevant ones. It reduces computational cost while preserving important attention interactions.

\textbf{Description:}  
The vertical-slash attention algorithm identifies the top-$k$ positions in vertical and diagonal slices of the attention matrix. It constructs a sparse attention index and computes the final attention output using these indices.

\begin{algorithm}
\caption{Efficient sparse attention computation by identifying the most relevant vertical and diagonal patterns in the attention matrix.}
\label{alg:vertical_slash}
\begin{algorithmic}[1]
\Require Queries $\mathbf{Q}$, Keys $\mathbf{K}$, Values $\mathbf{V} \in \mathbb{R}^{N \times d_h}$; vertical positions $k_v$, diagonal positions $k_s$
\Ensure Sparse attention output $\mathbf{y}$
\State \textbf{Compute Dense Attention Scores:}
\State $\mathbf{A} \gets \text{Softmax}(\mathbf{QK}^\top / \sqrt{d} + \mathbf{m}_{\text{causal}})$
\State \textbf{Select Top Indices:}
\State Top-$k$ vertical indices: $\mathbf{i}_v \gets \text{TopK}(\text{SumRows}(\mathbf{A}), k_v)$
\State Top-$k$ diagonal indices: $\mathbf{i}_s \gets \text{TopK}(\text{SumDiagonals}(\mathbf{A}), k_s)$
\State \textbf{Build Sparse Index:}
\State $\mathcal{I}_{\text{vs}} \gets \text{CombineIndices}(\mathbf{i}_v, \mathbf{i}_s)$
\State \textbf{Compute Sparse Attention:}
\State $\mathbf{A}_{\text{sparse}} \gets \text{SoftmaxSparse}(\mathbf{QK}^\top, \mathcal{I}_{\text{vs}}) / \sqrt{d}$
\State $\mathbf{y} \gets \text{SparseProduct}(\mathbf{A}_{\text{sparse}}, \mathbf{V}, \mathcal{I}_{\text{vs}})$
\State \Return $\mathbf{y}$
\end{algorithmic}
\end{algorithm}

\subsection{Block-Sparse Attention Algorithm}
\textbf{Algorithm Idea:}  
The block-sparse attention algorithm reduces complexity by dividing the attention matrix into fixed-size blocks and computing attention only for the most relevant ones. It leverages block-wise sparsity for efficient processing while retaining accuracy.

\textbf{Description:}  
This algorithm uses block-wise pooling to approximate the attention scores. It selects the top-$k$ most important blocks, constructs a sparse attention index, and computes the final output using these indices.

\begin{algorithm}
\caption{Optimized sparse attention by focusing computations on the most relevant blocks in the attention matrix.}
\label{alg:block_sparse}
\begin{algorithmic}[1]
\Require Queries $\mathbf{Q}$, Keys $\mathbf{K}$, Values $\mathbf{V} \in \mathbb{R}^{N \times d_h}$; block size $b$, number of blocks $k_b$
\Ensure Sparse attention output $\mathbf{y}$
\State \textbf{Compute Blocked Representation:}
\State Blocked queries: $\mathbf{Q}_{\text{block}} \gets \text{BlockMean}(\mathbf{Q}, b)$
\State Blocked keys: $\mathbf{K}_{\text{block}} \gets \text{BlockMean}(\mathbf{K}, b)$
\State Blocked attention scores: $\mathbf{A}_{\text{block}} \gets \text{Softmax}(\mathbf{Q}_{\text{block}} \mathbf{K}_{\text{block}}^\top / \sqrt{d} + \mathbf{m}_{\text{causal}})$
\State \textbf{Select Top Blocks:}
\State $\mathcal{I}_{\text{blocks}} \gets \text{TopKBlocks}(\mathbf{A}_{\text{block}}, k_b)$
\State \textbf{Compute Sparse Attention:}
\State Sparse attention scores: $\mathbf{A}_{\text{sparse}} \gets \text{SoftmaxSparse}(\mathbf{QK}^\top, \mathcal{I}_{\text{blocks}}) / \sqrt{d}$
\State Sparse output: $\mathbf{y} \gets \text{SparseProduct}(\mathbf{A}_{\text{sparse}}, \mathbf{V}, \mathcal{I}_{\text{blocks}})$
\State \Return $\mathbf{y}$
\end{algorithmic}
\end{algorithm}

\subsection{Main Idea}  
The main idea behind our approach is to leverage algorithms that enable highly efficient sparse computation for the attention matrix in long-context LLMs. The method is based on the following key steps:  

\begin{itemize}  
    \item \textbf{Categorization of Attention Heads:} The attention heads are categorized into three distinct types, allowing for targeted optimizations based on their specific characteristics.  

    \item \textbf{Kernel-Aware Sparse Pattern Selection:} Using a kernel-aware search method, we dynamically identify the optimal sparse pattern for each attention head, ensuring efficient computation with minimal overhead.  

    \item \textbf{Dynamic Sparse Masking:} A fast approximation approach is employed to construct dynamic sparse masks tailored to different input sequences. These masks are then applied to efficiently perform sparse attention calculations.  
\end{itemize}  

\subsection{Limitations and Trade-Offs}

While our method significantly reduces computational costs and latency for long input sequences, it introduces additional overhead for shorter inputs. This is because the extra computations required to determine the optimal sparse patterns and construct dynamic masks outweigh the performance benefits in such cases.

\subsection{Effectiveness Threshold}
Our primary goal is to accelerate model inference for relatively long contexts, ranging from 8K to 128K tokens, on mid-range hardware such as dual NVIDIA A5000 GPUs (24GB each). Therefore, an indirect objective of our approach is to minimize the threshold input sequence length at which our method becomes effective. For example, we aim to ensure that even for contexts as short as 16K tokens, the overall processing time, including the additional computations, is still lower than that of traditional dense attention methods. This threshold, where our method begins to deliver performance gains—measured via TTFT—demonstrates the practical utility of SparseAccelerate, even with the added computational steps.

\section{Experimental Results}  

We evaluate the performance of SparseAccelerate across various long-context benchmarks and real-world LLMs. Our experiments focus on three key aspects: latency reduction with an effectiveness threshold, memory efficiency, and accuracy retention.

We conducted experiments using the following configurations:  

\begin{itemize}  
    \item \textbf{Model:}  
    meta-llama/Llama-3.1-8B-Instruct\footnote{\url{https://huggingface.co/meta-llama/Llama-3.1-8B-Instruct}}, trained to handle up to 128M-token contexts.  

    \item \textbf{Attention Mechanisms:}  
    Comparative evaluations were performed across various attention mechanisms, including the default attention in LLaMA-3~\cite{grattafiori2024llama3herdmodels}, FlashAttention-2~\cite{dao2023flashattention2fasterattentionbetter}, SPDA~\footnote{\url{https://pytorch.org/docs/stable/generated/torch.nn.functional.scaled_dot_product_attention.html}}, Eager attention\cite{10.1145/3579371.3589057}, and our proposed SparseAccelerate.
    \item \textbf{Input Token Lengths:} Small , Medium, Large, Very Large Lengths.
    \item \textbf{GPU:} Dual NVIDIA A5000 GPUs, each with 24GB of memory, totaling 48GB. The Accelerate library\footnote{\url{https://huggingface.co/docs/accelerate/en/index}} is used for automatic device allocation and model parallelism across the 2 GPUs.

    \item \textbf{Metrics:}  
    The evaluation was based on the following metrics:  
    \begin{itemize}  
        \item Time-To-First-Token (TTFT): Measuring the latency to generate the first output token\footnote{\url{https://docs.nvidia.com/nim/benchmarking/llm/latest/metrics.html}}.  
        \item GPU Memory Usage: Assessing the memory requirements of different attention mechanisms.  
        \item Model Performance: Evaluating the accuracy of the models.  
    \end{itemize}  
\end{itemize}

\subsection{Latency Reduction and Effectiveness Threshold} 

\begin{table}[h!]
\centering
\begin{tabular}{c|ccccc}
\hline
\textbf{ctx} & \textbf{default} & \textbf{flashattention2} & \textbf{spda} & \textbf{eager} & \textbf{our} \\
\hline
10    & 0.4381  & 0.3672  & 0.3994  & 0.3988  & 2.9376  \\
50    & 0.4176  & 0.4035  & 0.3659  & 0.3873  & 3.6412  \\
100   & 0.6134  & 0.3764  & 0.3794  & 0.4583  & 3.8863  \\
500   & 0.4257  & 0.4464  & 0.4990  & 0.4506  & 4.2050  \\
1000  & 0.5319  & 0.5265  & 0.5443  & 0.5036  & 3.9370  \\
2000  & 0.8250  & 0.8339  & 0.7496  & 1.0656  & 4.2897  \\
4000  & 1.2269  & 1.3181  & 1.1606  & 2.4236  & 4.5298  \\
8000  & 2.2902  & 2.2521  & 2.3460  & -       & 5.8242  \\
16000 & 4.9286  & 4.7635  & 4.8810  & -       & 6.4833  \\
32000 & 11.6073 & 11.4526 & 11.5421 & -       & 11.1209 \\
64000 & -       & -       & -       & -       & 19.5932 \\
128000& -       & -       & -       & -       & 36.3091 \\
\hline
\end{tabular}
\caption{TTFT (Time to First Token) values for various attention mechanisms across different context lengths. A dash ('-') indicates an out-of-memory (OOM) condition.}
\label{table:result_ttft}
\end{table}

\begin{figure}[ht]
    \centering
    \includegraphics[width=0.8\textwidth]{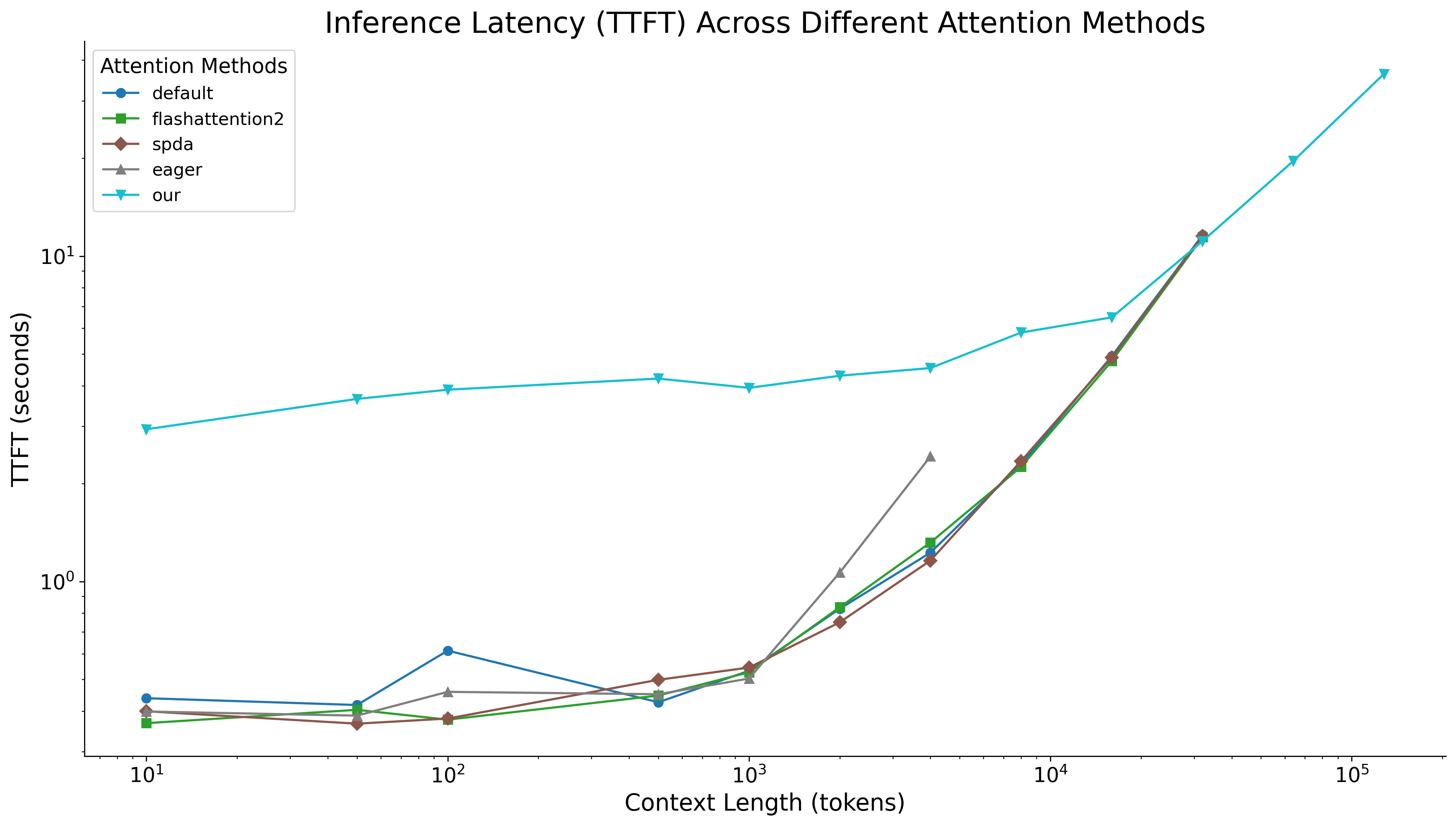}
    \caption{Inference Latency (TTFT) Across Different Attention Methods.}
    \label{fig:inference_latency}
\end{figure}

The experimental results for TTFT are presented in Table~\ref{table:result_ttft} and illustrated in Figure~\ref{fig:inference_latency}.

\paragraph{Small Context Lengths (10--100 Tokens)} 
At very small context sizes (10 tokens), most methods---default, flashattention2, spda, and eager---produce TTFT values well under one second (approximately 0.36--0.46 seconds). This indicates that these commonly used or optimized methods are highly efficient for short prompts. By contrast, our method begins with a much higher TTFT at 10 tokens (2.94 seconds), which is significantly slower than the others at this scale. While the standard methods excel at small contexts, our custom approach exhibits an initial overhead that is not present in the other methods.

\paragraph{Medium Context Lengths (500--4000 Tokens)} 
As we increase the context size from a few hundred to a few thousand tokens, default, flashattention2, and spda remain relatively competitive with one another. At 1000 tokens, their TTFT values cluster around 0.5--0.6 seconds, which remains quite low. Even at 4000 tokens, these methods hover around 1.0--1.3 seconds, indicating that they scale efficiently up to a few thousand tokens. The eager method is also efficient up to 4000 tokens, though we observe a larger increase in TTFT from 1000 to 4000 tokens: from about 1.07 seconds at 2000 tokens to 2.42 seconds at 4000 tokens. Although this is not exceedingly large, eager scales somewhat less smoothly than flashattention2 or spda in this range. 

Our method, while starting slowly, grows at a more modest rate as the context length increases. By 4000 tokens, its TTFT is approximately 4.53 seconds. Although this is slower than the sub-2-second times of the other methods at the same scale, its growth rate from 10 to 4000 tokens is more stable than one might expect given the high initial overhead.

\paragraph{Large Context Lengths (8000--32000 Tokens)} 
When we reach even larger context sizes, we observe more pronounced scaling effects. The default, flashattention2, and spda methods all exceed 2 seconds by 8000 tokens and surpass 4--5 seconds by 16,000 tokens. By 32,000 tokens, their TTFT values approach or exceed 11 seconds, indicating a significant increase in complexity at large context lengths.

Our method requires about 5.82 seconds at 8000 tokens, which, while higher than the others (e.g., default is 2.29 seconds and spda is 2.35 seconds at 8000 tokens), is not vastly greater. By 16,000 tokens, our method takes 6.48 seconds, whereas other methods are nearing or exceeding 4--5 seconds. By 32,000 tokens, our method’s TTFT is 11.12 seconds---remarkably close to the times of the other methods, all of which are around 11 seconds at this scale.

This convergence at large context sizes suggests that our method, initially disadvantaged at small scales, becomes competitive as context lengths grow very large. Its scaling curve appears more linear, or at least less rapidly accelerating, than those of the other methods at these larger scales.

\paragraph{Very Large Context Lengths (64,000--128,000 Tokens)} 
Only our method can successfully perform inference at these extremely large context lengths; all other methods encountered out-of-memory errors. At 64,000 tokens, our method reports a TTFT of 19.59 seconds, and at 128,000 tokens, 36.31 seconds. Since no other methods are tested beyond 32,000 tokens, there are no direct comparisons at these extreme scales. However, our method’s ability to operate at these very large contexts, and its approximately linear growth in TTFT, demonstrates considerable robustness and scalability.

\paragraph{Overall Performance Trade-Offs} 
For small to medium contexts (up to a few thousand tokens), established methods (default, flashattention2, spda, eager) achieve very low latency, making them ideal for scenarios where prompt lengths are moderate. In contrast, our method is not competitive at these smaller scales due to its high initial overhead.

For very large contexts (tens of thousands of tokens), the performance differences narrow. While our method starts slow, it scales in a manner that allows it to remain roughly in line with the others by the time we reach 32,000 tokens.

For extremely large contexts (beyond 32,000 tokens), only our method our method can successfully perform inference. It successfully scales to 128,000 tokens, maintaining a predictable increase in TTFT.

\paragraph{Effectiveness Threshold}
The experimental results indicate that the effectiveness threshold—where our method begins to deliver performance gains—is at 32K tokens, with a demonstrated improvement of up to 1.04x. In other words, our approach currently only becomes effective for input sequences exceeding 32K tokens. Lowering this threshold remains a key objective for future improvements.

\subsection{Memory Efficiency}  

\begin{table}[h!]
\centering
\begin{tabular}{c|ccccc}
\hline
\multirow{2}{*}{ctx} & \multicolumn{5}{c}{Attention Method} \\
 & default & flashattention2 & spda & eager & our \\
\hline
\multirow{3}{*}{10} 
 & 7161 & 7161 & 7161 & 7161 & 7161 \\ 
 & 8841 & 8841 & 8841 & 8841 & 8841 \\ 
 & 16002 & 16002 & 16002 & 16002 & 16002 \\ 
\hline

\multirow{3}{*}{50} 
 & 7177 & 7177 & 7177 & 7179 & 7163 \\ 
 & 8857 & 8857 & 8857 & 8857 & 8843 \\ 
 & 16034 & 16034 & 16034 & 16036 & 16006 \\ 
\hline

\multirow{3}{*}{100} 
 & 7193 & 7193 & 7193 & 7195 & 7167 \\ 
 & 8873 & 8873 & 8873 & 8875 & 8847 \\ 
 & 16066 & 16066 & 16066 & 16070 & 16014 \\ 
\hline

\multirow{3}{*}{500} 
 & 7337 & 7337 & 7337 & 7387 & 7215 \\ 
 & 9019 & 9019 & 9019 & 9071 & 8895 \\ 
 & 16356 & 16356 & 16356 & 16458 & 16110 \\ 
\hline

\multirow{3}{*}{1000} 
 & 7543 & 7543 & 7543 & 7831 & 7315 \\ 
 & 9225 & 9225 & 9225 & 9513 & 9003 \\ 
 & 16768 & 16768 & 16768 & 17344 & 16318 \\ 
\hline

\multirow{3}{*}{2000} 
 & 7959 & 7943 & 7959 & 9219 & 7473 \\ 
 & 9615 & 9599 & 9615 & 10413 & 9177 \\ 
 & 17574 & 17542 & 17574 & 19632 & 16650 \\ 
\hline

\multirow{3}{*}{4000} 
 & 8789 & 8757 & 8789 & 13283 & 7801 \\ 
 & 10441 & 10377 & 10441 & 14967 & 9527 \\ 
 & 19230 & 19134 & 19230 & 28250 & 17328 \\ 
\hline

\multirow{3}{*}{8000} 
 & 10425 & 10361 & 10425 & - & 8437 \\ 
 & 12029 & 11903 & 12029 & - & 10209 \\ 
 & 22454 & 22264 & 22454 & - & 18646 \\ 
\hline

\multirow{3}{*}{16000} 
 & 13695 & 13569 & 13695 & - & 9715 \\ 
 & 15245 & 14993 & 15245 & - & 11581 \\ 
 & 28940 & 28562 & 28940 & - & 21296 \\ 
\hline

\multirow{3}{*}{32000} 
 & 20259 & 20007 & 20259 & - & 12531 \\ 
 & 21625 & 21123 & 21625 & - & 14329 \\ 
 & 41884 & 41130 & 41884 & - & 26860 \\ 
\hline

\multirow{3}{*}{64000}
 & - & - & - & - & 14785 \\ 
 & - & - & - & - & 15209 \\ 
 & - & - & - & - & 29994 \\ 
\hline

\multirow{3}{*}{128000}
 & - & - & - & - & 17163 \\ 
 & - & - & - & - & 17589 \\ 
 & - & - & - & - & 34752 \\ 

\hline
\end{tabular}
\caption{GPU memory usage (GPU0, GPU1, and total in MB) for various attention methods at different context lengths (ctx). A dash ('-') indicates an out-of-memory (OOM) condition.}
\label{table:result_gpu}

\end{table}

\begin{figure}[ht]
    \centering
    \includegraphics[width=0.8\textwidth]{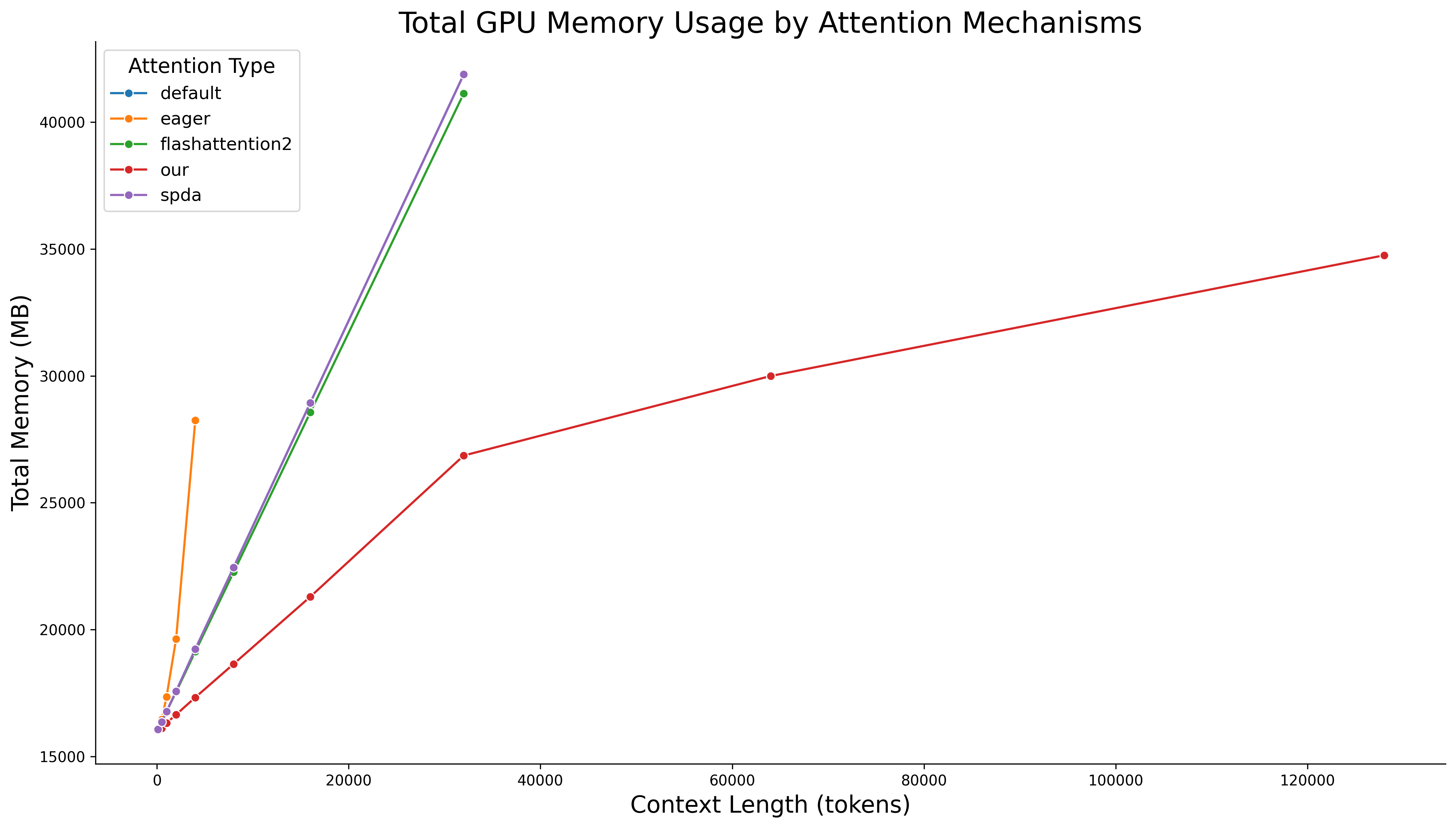}
    \caption{GPU memory usage Across Different Attention Methods.}
    \label{fig:inference_memory_plot}
\end{figure}

The observed results for GPU memory usage during inference are shown in in Table~\ref{table:result_gpu} and illustrated in Figure~\ref{fig:inference_memory_plot}.

\paragraph{Memory Scaling With Context Length:} Across all methods, GPU memory usage increases as the context length grows.

\paragraph{Small Context Lengths (10--100 Tokens)} At very small context lengths (10, 50, and 100 tokens), all attention methods—default, flashattention2, spda, eager, and our—consume nearly identical amounts of GPU memory. For instance, at 10 tokens, the total memory usage is 16,002 MB across all methods, while at 50 tokens, it remains roughly 16,006–16,036 MB for all methods. Even at 100 tokens, all methods remain in the same narrow range, with only a negligible advantage for the proposed ("our") method. This indicates that at minimal sequence lengths, the memory overhead is dominated by baseline model components rather than the specific attention implementations.

\paragraph{Medium Context Lengths (500--4000 Tokens)}
Starting from 500 tokens, differences in memory usage begin to emerge. While default, flashattention2, and spda remain closely aligned, the eager method starts using slightly more memory (e.g., at 500 tokens, eager totals 16,458 MB compared to ~16,356 MB for default/flash/spda). In contrast, our method uses distinctly less memory—about 16,110 MB at 500 tokens, which is a savings of a few hundred MB compared to other methods. These savings become more pronounced at higher medium-range contexts. For instance, at 2,000 tokens, eager’s memory usage spikes to 19,632 MB total, while our method uses only 16,650 MB—almost a 3,000 MB reduction. By the time we reach 4,000 tokens, eager runs out of memory (OOM), while default/flash/spda hover around 19,130–19,230 MB, and our method uses just 17,328 MB. This demonstrates that our approach scales more efficiently, allowing for larger sequence lengths without hitting memory limits.

\paragraph{Large Context Lengths (8000--32000 Tokens)}
At larger context lengths, such as 8,000 tokens, even more striking differences appear. Default, flashattention2, and spda can still run but consume around 22,264–22,454 MB. Eager fails to run at this scale, indicating a significant memory bottleneck. Meanwhile, our method uses only 18,646 MB—substantially less than other working methods. As context lengths increase to 16,000 and 32,000 tokens, the gap widens further. Default, flashattention2, and spda either approach or exceed 28,000 MB, while eager remains out-of-memory. Our method, by contrast, maintains a total GPU memory usage of about 21,296 MB at 16,000 tokens and 26,860 MB at 32,000 tokens. This marked difference highlights the superior memory scaling of our approach, enabling it to handle sequences twice or four times as long as other methods before running out of memory.

\paragraph{Very Large Context Lengths (64,000--128,000 Tokens)}
When scaling context lengths to 64,000 tokens and beyond, all methods except our approach are unable to run, leading to OOM conditions. At 64,000 tokens, our method uses about 29,994 MB total, and at 128,000 tokens, it uses 34,752 MB total. No other method can reach these extremely large sequence lengths without failing. This demonstrates that the memory efficiency of our attention mechanism is not merely incremental—it is transformative, allowing computations on sequences that are orders of magnitude longer than what other methods can handle.

\paragraph{Overall}
The data strongly suggests that our attention method achieves substantial memory savings over baseline methods, particularly as context lengths grow large. At small scales, differences are negligible, but at medium and large context lengths, our method consistently uses fewer MBs of GPU memory and can handle sequence lengths that cause other implementations to fail. Ultimately, this improved memory efficiency translates into the ability to handle much larger inputs, enabling practical use in memory-intensive applications and long-context tasks that were previously infeasible with standard attention mechanisms.

\subsection{Model Accuracy Preservation} 
This subsession is being updated.

\section{Conclusion}

In this paper, we introduced SparseAccelerate, a novel approach for reducing Time-To-First-Token (TTFT) latency in long-context LLM inference. Our primary focus is on enabling models to efficiently process contexts ranging from 8K to 128K tokens on mid-range hardware—specifically, dual NVIDIA A5000 GPUs with 24GB of memory each. An essential objective of our approach is to lower the threshold input sequence length at which our method becomes effective, ideally to as low as 16K tokens, ensuring that the total processing time (including additional computations) is still lower than that of traditional dense attention mechanisms.

SparseAccelerate leverages dynamic sparsity patterns and a kernel-aware optimization framework to address the quadratic scaling inherent in attention computations. Unlike static sparse methods, our approach dynamically adapts to input-specific attention distributions, allowing computational resources to be utilized more effectively. Experimental results demonstrate that while SparseAccelerate introduces an overhead at very short sequence lengths, it significantly outperforms competing methods as context lengths increase. Notably, it achieves up to a 1.04x reduction in TTFT latency at 64K-token prompts and maintains efficiency up to 128K tokens.

In addition to latency benefits, SparseAccelerate delivers substantial memory savings. Although differences are minimal at small scales, our method consistently uses fewer GPU resources at medium to large context lengths, enabling the model to handle longer input sequences without running out of memory. This memory efficiency translates into practical utility for memory-intensive applications and long-context tasks—previously infeasible with standard attention mechanisms.

By overcoming quadratic attention complexity, SparseAccelerate sets a new benchmark for scaling long-context LLMs on accessible hardware. Its ability to handle increasingly large inputs and achieve a lower effectiveness threshold empowers real-time, context-rich applications such as large-scale retrieval-augmented generation, long-form question answering, and advanced document comprehension. Looking ahead, we plan to explore additional sparsity patterns, refine heuristic search algorithms, and integrate with emerging acceleration frameworks, further reducing the effective threshold and broadening the range of practical, low-latency, large-context LLM deployments.

\bibliographystyle{plain}
\bibliography{references}

\end{document}